\documentclass{article}

\usepackage{arxiv}

\usepackage[utf8]{inputenc} % allow utf-8 input
\usepackage[T1]{fontenc}    % use 8-bit T1 fonts
\usepackage{hyperref}       % hyperlinks
\usepackage{url}            % simple URL typesetting
\usepackage{booktabs}       % professional-quality tables
\usepackage{amsfonts}       % blackboard math symbols
\usepackage{nicefrac}       % compact symbols for 1/2, etc.
\usepackage{microtype}      % microtypography
\usepackage{graphicx}
\usepackage{natbib}
\usepackage{doi}
\usepackage{algorithm}      % Float support
\usepackage{algpseudocode}  % Pseudocode commands
\usepackage{amsmath}

\title{A Hashgraph-Inspired Consensus Mechanism for Reliable Multi-Model Reasoning}

\author{ \href{https://orcid.org/0000-0001-5757-1101}{\includegraphics[scale=0.06]{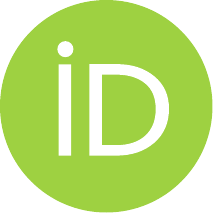}\hspace{1mm}Kolawole E. Ogunsina} \thanks{Corresponding author: kolawole08@gmail.com} \\
	\texttt{kolawole08@gmail.com} \\
	\And
	\href{https://orcid.org/0000-0002-7115-4651}{\includegraphics[scale=0.06]{orcid.pdf}\hspace{1mm}Morayo A. Ogunsina} \\
	\texttt{morayo.ogunsina@gmail.com} \\
}

\renewcommand{\shorttitle}{ Reliable Multi-Model Reasoning }

\hypersetup{
pdftitle={A Hashgraph-Inspired Consensus Mechanism for Reliable Multi-Model Reasoning},
pdfsubject={cs.AI},
pdfauthor={Kolawole E.~Ogunsina, Morayo A.~Ogunsina},
pdfkeywords={Decentralized AI validation, Multi-agent reasoning},
}

\begin{document}
\maketitle

\begin{abstract}
Inconsistent outputs and hallucinations from large language models (LLMs) are major obstacles to reliable AI systems. When different proprietary reasoning models (RMs), such as those by OpenAI, Google, Anthropic, DeepSeek, and xAI, are given the same complex request, they often produce divergent results due to variations in training and inference. This paper proposes a novel consensus mechanism, inspired by distributed ledger technology, to validate and converge these outputs, treating each RM as a black-box peer. Building on the Hashgraph consensus algorithm, our approach employs gossip-about-gossip communication and virtual voting to achieve agreement among an ensemble of RMs. The output of each model is gossiped through the network of models, and a Hashgraph-like protocol determines when each model has seen the output of all others. We then perform a virtual vote to identify a trusted result, minimizing hallucinations, and maximizing fidelity. Theoretical analysis shows that our mechanism retains blockchain-grade properties of consistency, fairness, and Byzantine fault tolerance, ensuring that even if some models produce misleading content, the ensemble can still agree on a correct answer. We present an architectural design for a prototype system in which RMs iteratively exchange and update their answers, using information from each round to improve accuracy and confidence in subsequent rounds. This approach goes beyond simple majority voting by incorporating the knowledge and cross-verification content of every model. We justify the feasibility of this Hashgraph-inspired consensus for AI ensembles and outline its advantages over traditional ensembling techniques in reducing nonfactual outputs. Preliminary considerations for implementation, evaluation criteria for convergence and accuracy, and potential challenges are discussed. The proposed mechanism demonstrates a promising direction for multi-agent AI systems to self-validate and deliver high-fidelity responses in complex tasks.
\end{abstract}

% keywords can be removed

\section{Introduction}
Hallucinations, outputs from AI models that are factually incorrect or fabricated, undermine trust in large language models (LLMs). LLMs often generate plausible but nonfactual content, making their reliability a concern \citep{Huang2023}. This problem is exacerbated in scenarios where multiple advanced reasoning models (RMs), each with proprietary architectures and training data, are available to handle the complex request of a user. In practice, one model may return an answer with certain details, while another provides a different explanation or even contradicts the first. This variability arises because each RM has unique internal reasoning and knowledge gaps. If these models are used independently, users will face inconsistent information and will not be able to easily discern the correct answer. A naïve solution is to take a majority vote or prefer the answer of a supposedly more accurate model. However, majority voting or static ensembling can fail when each model holds a piece of the truth or when the minority model is actually correct. Traditional consensus methods in machine learning assume an honest majority, whereas here the opinion of each model matters – even a single model can have crucial information that others missed \citep{Dursun2021}. We need a strategy that combines the strengths of all models while filtering out errors and hallucinations.

In distributed systems, consensus algorithms ensure that multiple agents (or nodes) agree on a single truth even in the presence of faulty or malicious participants. In particular, the Hashgraph consensus mechanism achieves high-efficiency asynchronous Byzantine fault tolerance by employing a gossip protocol and virtual voting
\citep{Baird2016swirlds}, \citep{Crary2021}. Inspired by this, our work treats each RM as a node in a distributed ledger-like network. We propose that RMs can collaboratively reach a consensus on their outputs for a given query, as much as the nodes on a blockchain agree on the next block. By adopting a Hashgraph-style consensus, we aim to leverage its fairness and fault tolerance to address AI ensemble output verification. In essence, the RMs will gossip their answers to each other and virtually vote on the correctness of information, iteratively refining their responses until convergence.

\textbf{Contributions}: This paper introduces a Hashgraph-inspired consensus mechanism for multi-model AI reasoning and provides both theoretical underpinnings and a prototype design. Key contributions include:

\begin{itemize}
    \item \textbf{\textit{Consensus Mechanism Design}}: We detail how gossip about gossip and virtual voting can be adapted to propagate and validate information among black-box RMs. Our design ensures that if any model discovers a correct piece of information, it spreads exponentially to all other models \citep{Hedera2025gossip}, while conflicting or hallucinated content is identified and pruned by voting.
    \item \textbf{\textit{Iterative Convergence Protocol}}: We propose an iterative protocol where each RM updates its response after receiving the output of the others. We define criteria for completing a consensus round (each model having seen the information from all others) and for deciding final convergence. Unlike a one-shot majority vote, this multi-round deliberation enables collective reasoning until the models reach unanimous or high-confidence agreement.
    
    \item \textbf{\textit{Prototype Architecture}}: We outline an architecture to implement the consensus mechanism, treating proprietary RMs as services and using an external coordinator or peer-to-peer messaging for gossip. The design describes how to encode the results as transactions in a Hashgraph-like data structure and how the local view of each RM leads to a consistent global decision without centralized control.
    
    \item \textbf{\textit{Evaluation Framework}}: We discuss how to evaluate the effectiveness of the consensus mechanism in reducing hallucinations and improving accuracy. We identify metrics such as the factual accuracy of the final answer, the rate of hallucinated content, the number of rounds to convergence, and robustness against malicious or outlier models.
\end{itemize}

This approach, to our knowledge, is one of the first to apply distributed ledger consensus principles to multi-agent AI validation. By combining ideas from blockchain consensus and LLM ensemble techniques, it offers a novel way to increase reliability in AI systems. The rest of the paper is organized as follows: we formally define the problem and review related work in Section \ref{sec:probdef} and Section \ref{sec:relwork}, respectively. In Section \ref{sec:solution}, we present the theory and architecture of our consensus mechanism. Section \ref{sec:design} describes a potential implementation in detail. We then outline the evaluation criteria in Section \ref{sec:eval} and discuss the implementation challenges in Section \ref{sec:challenge}. Finally, we summarize our findings and future prospects in Section \ref{sec:conclusion}.

\section{Problem Definition} \label{sec:probdef}
\textbf{Ensemble of Divergent AI Outputs}: Consider a set of $N$ independent reasoning models ${RM_1, RM_2, \dots, RM_N}$, each provided with the same complex user query or task. Each $RM_i$ produces an output $O_i$ (e.g., an answer to a question, a piece of code, or a step-by-step reasoning). Because these models have been trained on different data and have different inference heuristics, the results ${O_1, O_2, \dots, O_N}$ may vary significantly in content, format, and correctness. Our focus is on complex queries that require multistep reasoning, factual knowledge retrieval, or adherence to detailed instructions; tasks where large models sometimes hallucinate facts or stray from the instructions. A hallucination is defined as any content generated by $RM_i$ that is not supported by factual data or the prompt, often sounding plausible but incorrect \citep{Huang2023}. Divergent outputs mean that the ensemble of RMs does not automatically agree on a single result, leaving ambiguity about the truth.

\textbf{Goals}: We seek a mechanism to aggregate these $N$ outputs into a single consensus output $O^{*}$ that: 
\begin{itemize}
    \item has higher factual accuracy and faithfulness to the user’s request than any individual $O_i$; 
    
    \item is agreed upon (identical or equivalently endorsed) by all RMs after a consensus process. 
\end{itemize}
Importantly, the process should minimize the inclusion of hallucinated or erroneous content in $O^{*}$. In other words, if some RMs produce nonfactual statements, the consensus mechanism should detect and exclude those, favoring content that withstands cross-model verification. Conversely, if a single RM contains a crucial correct detail that others omitted, the mechanism should ideally incorporate that detail into the final answer rather than lose it to a majority vote. This is a key difference from simplistic voting; we want to preserve valuable minority contributions while filtering out mistakes.

\textbf{Assumptions}: Each $RM_i$ is treated as a black box, which means that we cannot alter its internal weights or directly inspect its hidden reasoning. Our only interface is through inputs (queries or prompts) and outputs (the generated responses). We assume that we can prompt each RM multiple times if needed, including feeding it information (such as the output of other models) in subsequent rounds. This allows RMs to assess and respond to each other’s responses as if they were participating in a discussion. We also assume that a communication channel exists between models (directly or via an orchestrator) so that outputs can be exchanged. In distributed systems terms, each RM is a node that can send and receive messages (the content of outputs) to/from other nodes. We do not assume that any RM is inherently trusted over others; any of them could hallucinate or be \textit{Byzantine} (arbitrarily faulty) in their output. However, we assume that not all RMs will collude to mislead; at least a majority will act in good faith and have some correct grounding, which is analogous to the standard Byzantine fault tolerance assumption (Hashgraph tolerates up to 1/3 malicious nodes \citep{Baird2016swirlds}).

\textbf{Challenges}: The problem is challenging because natural language outputs are not simple boolean values or numbers that are easy to vote on; they are complex pieces of text or reasoning chains. Determining whether two outputs agree or conflict can itself be nontrivial. For example, the answer of a RM may contain additional details or a different explanation that is not obviously in conflict with the answer of another RM. We must detect subtle contradictions or hallucinations. Another challenge is fidelity to the user request. The consensus process must ensure that the final answer still addresses the query correctly and does not dilute the clarity by trying to appease all models. Additionally, the mechanism should converge in a reasonable number of iterations; it should detect when consensus is reached without unnecessary cycles. Finally, treating models as black boxes means that we rely on their ability to process new information in prompts. If a model is unwilling or unable to revise its answer when presented with the answers of its peers, consensus might never be reached. Our design must account for or mitigate such stubborn behavior (for example, by explicitly instructing models to consider the output of others as new evidence).

In summary, our problem is to design a consensus mechanism $M$ that takes ${O_1,\dots,O_N}$ as input (initial model outputs) and produces a final output $O^{*}$ with significantly reduced hallucinations and improved accuracy, under the constraints of black-box models and potential faults in some outputs. We aim for $O^{*}$ to be correct (or at least more correct than any $O_i$), consistent (agreed by all models after the process), and produced efficiently.

\section{Related Work} \label{sec:relwork}
\textbf{Hallucination Mitigation in LLMs}: A growing body of research focuses on detecting and reducing hallucinations in single-model output. \textit{SelfCheckGPT} \citep{morgan2025selfcheckgpt}, for example, uses consistency checks by prompting the same model multiple times and verifying if the answers align. If a fact is true, repeated sampled responses from one model should agree, whereas hallucinations often lead to divergent answers \citep{morgan2025selfcheckgpt}. This approach resembles an ensemble of outputs from one model. It employs an implicit consensus: if the model’s own \textit{jury} of answers disagrees, the content is flagged as possibly false. Although effective for single-model use, \textit{SelfCheckGPT} does not leverage the diversity of knowledge across different model architectures. Another technique is LLM-as-a-Judge \citep{Zheng2023}, where one model (or a smaller verifier model) is used to evaluate the output of another. These LLM juries extend this idea by having multiple models vote on the quality or correctness of an answer. These methods treat models as independent evaluators, but typically the final decision is still by majority or a weighted scheme, and there is no multi-turn deliberation among the models.

\textbf{Ensemble and Voting Methods}: Traditional ensemble learning in machine learning suggests that combining multiple models can improve overall performance and robustness. Techniques like majority voting, averaging, or taking the most confident prediction are common in classification tasks. For LLMs and generative models, a simple ensemble might involve generating multiple outputs and choosing the most frequent answer; a method used in self-consistency decoding for chain-of-thought reasoning \citep{Wang2022}. \cite{Wang2022} demonstrated that sampling multiple reasoning paths and then selecting the most consistent answer significantly increases accuracy on reasoning benchmarks. This self-consistency method is essentially a majority vote on the different reasoning realizations of one model. By analogy, different models could be ensembled by seeing which answer appears most often among them. However, majority voting in isolation has limitations: if the majority of models share the same blind spot or training bias, they may all hallucinate the same incorrect fact, leading the ensemble to confidently choose a wrong answer. In contrast, a single correct answer from a minority model could be discarded. Our approach seeks to avoid these pitfalls by enabling information exchange between models, rather than just a vote tally. In other words, we introduce communication and convergence, not just aggregation.

\textbf{Distributed Consensus Algorithms}: In distributed computing, consensus algorithms such as Paxos \citep{Lamport2001paxos}, Raft \citep{Ongaro2014raft}, and Byzantine Fault Tolerance (BFT) protocols such as PBFT \citep{Castro1999paxos} are designed to achieve agreement on a value among distributed processes or nodes, even if some fraction of nodes are faulty. Blockchain systems introduced variants such as Proof-of-Work and Proof-of-Stake to achieve consensus in open networks, but those rely on cryptographic puzzles or economic incentives rather than agreement on content. Hashgraph, introduced by \cite{Baird2016swirlds}, is a notable consensus algorithm that uses a directed acyclic graph (DAG) of gossip events rather than a linear chain of blocks. In Hashgraph, each node rapidly gossips information about transactions to others (gossip protocol) and records a history of who told whom (this history is the \textit{gossip-about-gossip} DAG) \citep{Baird2016swirlds}. Using this DAG, the nodes perform virtual voting to agree on the order of transactions without exchanging actual vote messages \citep{Baird2016swirlds}. The outcome is a consensus that is fair (i.e. no leader or miner can dictate the order) and fast. Key properties of Hashgraph relevant to our work include: (1) Exponential information propagation; new data reaches all nodes quickly, (2) Byzantine fault tolerance; the network can tolerate up to $1/3$ faulty nodes and still reach agreement, and (3) Deterministic agreement; all honest nodes calculate the same consensus outcome (e.g., same transaction order or, in our case, same agreed output). We draw inspiration from these properties to design a consensus among RMs (where the \textit{faulty} nodes are models that produce hallucinations or errors). In particular, gossip about gossip from Hashgraph ensures that every node knows what every other node has learned in the process, and virtual voting provides an elegant way to simulate majority agreement without additional communication overhead \citep{Crary2021}. In our adaptation, we will not literally compute cryptographic hashes or timestamps, but the conceptual framework of rounds of gossip and implicit voting guides our protocol.

\textbf{Multi-Agent Debate and Deliberation}: Another strand of related work is the idea of having multiple AI agents engage in a debate or deliberative dialogue to reach a correct answer. OpenAI’s AI safety via debate \citep{Irving2018debate} had two models argue opposing sides of an answer with a judge (which could be a human or another model) determining the winner. While debate is adversarial and aimed at exposing flaws (thus potentially uncovering truths), it requires a judging component and structured argumentation. \cite{Ogunsina2022} provided a comprehensive review of fusion techniques for AI and Hashgraph, detailing how their integration facilitated the development of a decentralized AI platform that simulates the integration and interaction of intelligent agents for simultaneously-integrated recovery during airline disruption management. They demonstrated the efficacy of the synthesis of AI and the Hashgraph consensus algorithm for airline operations recovery \citep{Ogunsina2022}. More recently, \cite{Pokharel2025} introduced a deliberation-based consensus mechanism in which multiple LLM agents discuss to reach unanimous decisions. They leverage graded consensus and multi-round deliberation to ensure that for clear questions the agents reach full agreement, while for subjective matters they output a distribution of opinions. Their approach maintains classic blockchain consensus properties (consistency, fairness, etc.) and explicitly tackles issues like degeneration of thoughts, hallucinations, and even malicious models during the multi-agent conversation \citep{Pokharel2025}. Our work is aligned in spirit with such multi-round deliberation. However, we center our design on a specific proven consensus protocol (Hashgraph) as the backbone, giving a clear rule-set for information exchange and decision making. This provides a rigorous theoretical foundation and the potential to formally prove guarantees about the outcomes (e.g., if at least 2/3 of models have a fact correct, the consensus will reflect that fact with high confidence). Furthermore, while \cite{Pokharel2025} focus on blockchain decision-making scenarios, our aim is specifically to reduce factual errors (hallucinations) in question-answering (QA) or task-solving contexts.

In summary, previous work provides valuable ideas: the use of multiple LLMs for cross-checking, the benefits of iterative dialogues among models, and the rich literature of distributed consensus for reliability. Our contribution is to integrate these ideas into a cohesive system where multiple black-box RMs effectively act as validators for each other, using a Hashgraph-inspired protocol to agree on a final answer. This goes beyond voting or simple ensembling by introducing a structured consensus mechanism for AI reasoning.

\section{Proposed Solution} \label{sec:solution}
Our proposed solution is a Hashgraph-inspired consensus mechanism for an ensemble of reasoning models. We first overview the approach, then detail the theoretical consensus algorithm and the architectural design in tandem.

\subsection{Overview of the Consensus Approach}
At a high level, we set up an iterative process in which all RMs share their outputs with each other and collectively decide on a final output. The process is analogous to the nodes reaching consensus on a transaction ledger, but here the \textit{transaction} is the correct answer to the user’s query. We divide the process into rounds. Round $0$ is the initial query-answer phase: each $RM_i$ independently generates an output $O_i^{0}$. Then the consensus rounds begin (Round $1$, Round $2$, \dots, Round $r$ ). In each round: 

\begin{itemize}
    \item \textbf{\textit{Gossip Phase}}: Models exchange information about their current answers (responses). Rather than a trivial broadcast, we envision using \textit{gossip about gossip} to manage this exchange. For example, $RM_1$ might randomly pick $RM_2$ and send its output $O_1$ to $RM_2$. $RM_2$ then combines that with its own information and gossips to another model, and so on \citep{Baird2016swirlds}. In practice, to ensure efficiency given typically small $N$, this gossip can be accelerated to ensure that every model receives the output of every other model by the end of the phase. The \textit{gossip-about-gossip} aspect means that each message can carry not only the original query result, but also metadata such as from which model the information came from and in which round, allowing recipients to track who knows what. This builds a shared history (a conceptual DAG of information exchange similar to a hashgraph) that all models incrementally learn. The gossip phase of a round is complete when each RM has received the latest output from all other RMs. At that point, every model has the same set ${O_1^{r-1}, ..., O_N^{r-1}}$ of outputs from the previous round.

    \item \textbf{\textit{Local Computation \& Virtual Voting}}: Once a model has the full set of peer outputs from the last round, it evaluates them to decide on its next output. This is where we implement a virtual voting mechanism. Rather than asking each model to explicitly vote on which answer is best (which would require additional queries or messages), we have each model internally decide how to update its answer given what it learned. We craft a prompt or algorithm that instructs $RM_i$: “\textit{Given the following answers from other models and your own previous answer, determine the most accurate and consensus answer.}”. Essentially, each model simulates being a voter that considers all candidates (answers) and then produces a new answer, which can be seen as its vote for the truth. Because every model received the same information, if they follow a similar decision rule, we expect their new outputs ${O_1^r, ..., O_N^r}$ to start aligning. This mirrors Hashgraph’s virtual voting, where each node independently computes votes based on its copy of the gossip graph. In our case, the \textit{vote} could be implicit: if most models saw that a particular fact or conclusion was present in the majority of peers, they are likely to include or agree with that in their next output (a form of majority vote occurring within the reasoning of each model). More formally, we can imagine that each $RM_i$ performs a function $f$ on the set of last-round outputs: $O_i^r = f(O_1^{r-1}, ..., O_N^{r-1})$. The goal is for the function $f$ (when applied by all models on the same input set) to produce agreement, i.e. $O_1^r \approx O_2^r \approx \dots \approx O_N^{r} = O^{*}$. This would be a fixed point (consensus) of the iteration.

    \item \textbf{\textit{Convergence Check}}: After each round’s local computation, we check if consensus is reached. Consensus can be defined as all models producing the same output (or semantically equivalent outputs) in round $r$, that is, $O_1^r = O_2^r = \cdots = O_N^r$. In practice, since the outputs are text, we define a tolerance (they do not have to be word-for-word identical if they only differ in phrasing). If consensus is reached, the process ends and that agreed output is taken as the final answer. If not, another round begins: the new outputs are gossiped, and models will update again. In Hashgraph terms, the process of gossip until all know everything and virtual voting in rounds continues until a famous witness or sufficient agreement emerges to finalize a decision \citep{Baird2016swirlds}. Here, the equivalent of deciding on a famous witness is that an answer has effectively won the support of a supermajority of models and becomes stable.
\end{itemize}

Using this approach, even if initially the models strongly disagree, through a few rounds of information exchange they may converge. Intuitively, this happens because correct information tends to be corroborated by multiple models (or at least not contradicted), while hallucinations or errors made by one model will be recognized by others; they might not repeat that claim if they know it is dubious or they will notice that it was not present in others’ answers and question it. Over rounds, unsupported claims should vanish (no one else repeats them and the original model might drop it if it sees that no one else had that idea, mimicking how a false rumor might die out in a consensus process), and well-supported claims will propagate (if several models had the same correct fact, even a model that missed it initially may adopt it after seeing it in peers’ answers). This is analogous to finding common truth through gossip: true facts reinforce each other, isolated errors get isolated and removed.

\subsection{Theoretical Underpinnings: Gossip and Virtual Voting for RMs}
We now draw parallels to Hashgraph more explicitly to frame the theoretical robustness of our mechanism:

\begin{itemize}
    \item \textbf{\textit{Gossip about Gossip}}: In Hashgraph, \textit{gossip about gossip} ensures that every node eventually knows the exact history of message spreading \citep{Hedera2025gossip}. For our purposes, we do not need to record an exact DAG of events with cryptographic hashes, but conceptually we maintain a log of which RM said what in round $p$ that every RM will eventually know. When $RM_i$ shares $O_i^{r-1}$ with $RM_j$, both $RM_i$ and $RM_j$ can note that $RM_i$ told $RM_j$ this. As gossip continues, all models can attain a shared view such as: $RM_1$ had answer A, $RM_2$ had answer B, etc. in round $0$. By round $1$, $RM_1$ has seen B and C (others' answers), $RM_2$ has seen A and C, and so on. This record helps in two ways. First, it guarantees transparency: no model can hide information. If one model found a certain solution, eventually all models will know that the solution was proposed. Second, it helps in knowing when to stop gossiping in a round: the round ends when the knowledge of the gossip graph is saturated, that is, each answer in round $r-1$ is present at every node. In Hashgraph terms, each gossip event becomes known to all RMs, making events \textit{ancestors} to all current events \citep{Hedera2025gossip}. We can formally define a round in our mechanism similar to the Hashgraph division into rounds \citep{Hedera2025virtualvoting}: start round $r$ when each model's knowledge of the results of round $r-1$ is complete. We could even define a notion of strongly seeing, such that $RM_i$ strongly sees $O_j^{r-1}$ if $RM_i$ not only received $O_j^{r-1}$ but knows that the majority of other models have also received it (this could be inferred via meta-information in gossip messages). This ensures that by the time we move to voting, the information is globally disseminated.
    
    \item \textbf{\textit{Virtual Voting and Fault Tolerance}}: Once all outputs are shared, deciding the \textit{consensus output} can be thought of as a voting problem: the current answer of each model is like a vote for a particular proposition (the content of the answer). Instead of holding an actual vote count externally, we rely on the models to incorporate others’ answers and effectively \textit{vote} with their updated answer. This is analogous to Hashgraph’s virtual voting where each node deduces votes from its local copy of the hashgraph \citep{Hedera2025virtualvoting}. The benefit is that we do not require additional communication, which would be akin to asking each model “do you agree with answer $X$?” explicitly. The local update function $f$ each model applies is critical: it should be designed such that if a large fraction of models have a particular piece of information in their answers, that information gets carried into the next round by all. For example, if four out of five models stated a specific fact $F$ and the fifth did not mention $F$, then in the next round, ideally all five models will include $F$ because seeing it in the answers of four peers signals it is likely correct. This is similar to a supermajority vote of $4/5$ in favor of $F$, which should convince the lone holdout. Hashgraph ensures that any value that is supported by over $2/3$ nodes becomes known to all honest nodes (Byzantine agreement) \citep{Hedera2025virtualvoting}. In our system, if at least $2/3$ of models state an answer or sub-answer, our update will propagate that to all models in the next round, achieving an analogous guarantee. Conversely, if a model produces a unique hallucinated detail that no other model has, no other model will include it, and ideally the original model will drop it once it realizes no peer validates it. This is analogous to a vote where only one node votes \textit{yes} and all others \textit{no} for a fact; the one \textit{yes} vote is outweighed, and the node should change its vote to \textit{no} in the next round.
    
    \item \textit{\textbf{Consensus Finalization}}: One nuance is deciding when the final answer is stable. In Hashgraph, once events are known and enough rounds of virtual voting occur, certain events (transactions) can be assigned a consensus order and timestamp with confidence (they become \textit{famous witnesses} in consensus) \citep{Hedera2025virtualvoting}. In our scenario, we do not have an ongoing stream of transactions, just the answer to one question. We can consider the final answer determined when an output has fame. That is, all models accept that it is the agreed answer. One can implement a condition like: if in round $r$ all models’ outputs match, or if the differences are only superficial, then consensus is reached. Additionally, we may require that this agreed answer persists for one additional round (to ensure stability). For example, if all models output identical $O^r$, we run one more round where they exchange $O^r$ (which is the same for everyone) and see that nothing changes in $O^{r+1}$. This is similar to a finality confirmation. If a slight difference remains (say two models phrased the same content differently), a simple heuristic could be applied to unify phrasing or pick the wording of one model as canonical, since they semantically agree.

\end{itemize}
\textbf{Why Hashgraph?}: The reason for adopting Hashgraph elements instead of a simpler approach (like just one round of sharing and a vote) is to inherit its strong guarantees. Hashgraph’s approach ensures that consensus is not only reached, but is correct under assumptions of a bounded fraction of bad actors. In our adaptation, \textit{bad actors} equate to models consistently providing wrong information. If, say, out of five models, one or two are hallucinating wildly or have outdated knowledge, our consensus mechanism should still converge to the truth as long as the other three or four have or can reason out the correct answer. This is analogous to requiring a supermajority of honest nodes for Byzantine consensus \citep{Baird2016swirlds}. Also, Hashgraph’s fairness (the information from no single node is preferred just by origin) is mirrored in our design; the input from every model is considered and propagated. We do not designate a leader model or always trust Model A over Model B. This removes biases and exploits the full knowledge of the ensemble. By using a random gossip process, we also avoid any fixed ordering in which models influence others, reducing the chance that a particular model always dominates the consensus. Each model ends up seeing a blended view of all answers, without knowing which model was the original source of a piece of information by the time it is fully gossiped (since gossip randomizes the pathway). This can prevent, for example, a bias where some might trust the model from OpenAI more. In our protocol, all models only see content, not identities.

\subsection{Prototype Architectural Details}\label{protodets}
To ground the theoretical design, here we describe how one could implement a prototype system that realizes this consensus mechanism. The system consists of the following components:

\begin{itemize}
    \item \textbf{\textit{Reasoning Model Pool}}: A set of $N$ black-box model endpoints (via APIs or local instances), e.g., \textit{OpenAI o3}, \textit{Anthropic Claude 3.7}, \textit{DeepSeek R1}, \textit{xAI Grok}, etc. Each can be queried with a prompt and will return a completion (text result). For the prototype, these models are treated as services we can call sequentially or in parallel.
    
    \item \textbf{\textit{Orchestrator / Mediator}}: A central controller that simulates the networking and consensus logic. In a fully decentralized setting, each model would need to host a consensus client to handle gossip and voting. In our prototype, we simplify by having one orchestrator process that queries each model and passes messages. This orchestrator maintains the gossip state. Essentially, a matrix of which model’s output each model has seen at each round. It triggers rounds and keeps track of outputs.
    
    \item \textbf{\textit{Round Execution}}: The orchestrator starts by sending the query from the user to all RMs in parallel to get initial outputs $O_i^0$. Then for round $1$ to $r$ (until convergence):
    \begin{itemize} 
        \item It executes the gossip phase. This can be done in simulated steps: pick random pairs or sequences to send outputs. For example, it may send the output from $RM_1$ to $RM_2$ (by constructing a prompt for $RM_2$ that says: \textit{“Another AI answered: [text of $O_1$]. Here is the question again: [question]. What is your answer?”}). Then send $RM_2$’s (now possibly updated or initial) response to $RM_3$ in a similar prompt, and so on. However, a more straightforward implementation approach is to compile a summary of all known outputs and present that to each model. By the end of round $1$ gossip, we want each model to have seen all $O_j^0, j \neq i$. In practice, we can achieve this in one step by prompting each model with a list of the answers from all other models. This is effectively one round of full information exchange rather than iterative random gossip to save time. The prompt template for $RM_i$ in round $1$ can be:
        
        \textit{"You are one of several AI agents solving a problem. The problem is: [user query]. Here are answers from your peers:
        \begin{itemize}
            \item $RM_1$: [$O_1^0$]
            \item $RM_2$: [$O_2^0$]
            \item (excluding $RM_i$’s own initial answer)
            Reconsider your answer in light of the above and provide a revised answer that you believe is most accurate and addresses the query faithfully.
        \end{itemize}
        "}
        
        This prompt explicitly gives $RM_i$ knowledge of all the answers from all other models. This is equivalent to an all-to-all gossip in one go. While this loses the step-by-step DAG buildup of gossip about gossip, the orchestrator can include tags like \textit{“$RM_1$ said ...”} to keep track of source; effectively it is a complete gossip sync where everyone receives everything \citep{Hedera2025gossip}. After sending this to each model, we receive updated answers $O_i^1$.
        \item The orchestrator then compares $O_1^1, O_2^1, ..., O_N^1$. If they are not yet the same, it proceeds to round $2$. For round $2$, the same procedure is applied: each model is prompted with the list of all outputs from round $1$ (including possibly its own, to show what everyone said). They each return $O_i^2$. We check for consensus, and so on.
    \end{itemize}
    This design is straightforward, though possibly communication-heavy (each round involves querying all models with a prompt containing the outputs from all models from the previous round). We can optimize by only including differences or a merged summary if needed.

    \item \textbf{\textit{Convergence and Termination}}: We need a method to decide that the outputs converged. The orchestrator can use a simple text similarity or exact match check. This is trivial for exact string match. If not exact match, we can use a semantic similarity metric or even ask another AI to verify if the answers are equivalent. For example, if two answers differ in wording but convey the same information, the orchestrator may deem that the consensus is acceptable. For a simple prototype, we can require exact or very high similarity to declare success. Once converged, the orchestrator returns the final answer $O^{*}$, which could be any of the outputs of the model from the last round, since they are now identical or we choose one representative.
\end{itemize}

This architecture ensures that, at each round, every model receives the knowledge of all others (fulfilling the consensus round condition) and that each model actively participates in refining the answer. Essentially, the orchestrator is implementing a synchronous version of the consensus. In a real distributed setup, models may gossip asynchronously, but here we simulate round-by-round synchrony for simplicity.

An important design detail is the content of the prompt for updates. We must encourage the model not to blindly copy the answer from another model, but to critically evaluate and integrate. For example, if the answer from one model contradicts another model on a factual point, the model should flag that and decide which is correct (or express uncertainty). We can modify the prompt as such: \textit{“Compare the following answers from your peers with yours. Identify any conflicting information. If most of your peers agree on something you missed, consider if they might be correct. If you stated something no one else did and you are not confident it is correct, you may remove or verify it. Now provide the answer you think is most likely correct.”}. This kind of instruction pushes the model toward consensus behavior: align with the group on supported facts, drop unsupported claims, and keep the answer coherent. Designing and fine-tuning these prompts is part of the prototype development to ensure that the concept of virtual voting translates into the reasoning of each model.

\textbf{Information from Each Round}: We maintain state across rounds. For example, a model might not reintroduce a fact it dropped earlier unless new evidence appears. In the orchestrator, we could even keep track of which model contributed which facts. Over rounds, we expect convergence, but the history of how they converged could be used to estimate confidence. For example, if in round $1$ there was wide disagreement and by round $3$ they agreed, we could use the number of rounds or the change in content as a signal of how contentious the query (question) was. Fewer rounds to converge may indicate a straightforward query (high confidence in final answer), whereas multiple rounds with many changes indicates the ensemble had to resolve conflicts. So, the final answer, while agreed upon, had to override some initial disagreements. This can be treated with a bit more caution requiring the models to provide explanations. Thus, the system can output not just the answer but a consensus report: e.g.,\textit{“All four models agreed on this answer after 2 rounds of discussion.”}. That is valuable to the end-user as a confidence indicator.

Another use of round information is iterative improvement beyond a single query. If we imagine a persistent system that handles many queries, models could adapt (or we could learn weightings) from past rounds. For example, if we notice one model frequently changes its answer to match others (indicating it often had mistakes initially), we can down-weight its independent answers in the future or expedite consensus by initially biasing toward the historically most reliable model. However, since we treat them as black boxes, such adaptive weighting would be outside the strict consensus protocol and edges into heuristic ensemble weighting. Our core mechanism remains unbiased, but we note this as a possible improvement: to learn which models are more often \textit{right} versus \textit{hallucinating} and incorporate that in the consensus process (similar to validator reputation in some consensus systems).

\subsection{Example Walk-through (Conceptual)}
\textit{For illustration, consider a simple factual question: \textit{“What is the capital of Australia?”}}
\begin{itemize}
    \item Suppose $RM_1$ initially answers \textit{“Sydney”} (a common mistake or hallucination), $RM_2$ answers \textit{“Canberra”} (correct), $RM_3$ answers \textit{“Canberra”}, and $RM_4$ answers \textit{“Canberra”}. Out of the four models, three have the correct answer and one is wrong. A majority vote would already pick \textit{“Canberra”}. In our consensus round $1$: each model sees others’ answers. $RM_1$ sees that three peers said \textit{“Canberra”} while only it said \textit{“Sydney”}. Recognizing it is likely wrong, $RM_1$ changes to \textit{“Canberra”}. The others see one outlier said \textit{“Sydney”} but most said \textit{“Canberra”}, and perhaps they themselves already had \textit{“Canberra”}, so they stick with \textit{“Canberra”}. In round $1$ outputs, all four now say \textit{“Canberra”}. Consensus achieved quickly; the orchestrator stops. This trivial example shows agreement in one round. The hallucination \textit{“Sydney”} was eliminated because it did not have support.
\end{itemize}
\textit{Now consider a more complex query requiring a multi-sentence answer, where each model provides overlapping but differing details, and one model injects a false detail.}
\begin{itemize}
    \item Initially, the answers vary: some have details $A$, $B$, $C$ (some correct, some not). After gossip, each model sees all details. Those details that appear in multiple answers (say $A$ and $B$ appear in the output of three models) will be regarded as credible by the others and included. The false detail $D$ that came from only one model will likely be dropped by that model when it sees no one else had $D$ (especially if the prompt encourages skepticism for unique claims). By round $2$, all models may include $A$ and $B$ in their answers, and omit $D$. If one model had a unique but correct detail $E$, it is trickier. Initially only it has $E$, others do not. Others may omit $E$ in round $1$. But the model with $E$ sees no one else had it. If it is confident (maybe it was a factual memory), it will keep $E$, but now it is effectively a minority of one. In round $2$, others see that this model still mentions $E$ (and now everything else they all agree on). If $E$ does not contradict anything and the ensemble prompt encourages considering all peer contributions, the others may start including $E$ in round $2$, especially if no one can refute $E$. After enough rounds, $E$ would propagate. This shows the system can amplify a true detail from minority to majority if it consistently remains and faces no opposition. In practice, whether this happens depends on model behavior (this is where perhaps an external fact-check or a rule could help if models themselves are not sure about $E$). However, the design encourages convergence: either the lone model drops $E$ (thinking it was wrong or not needed) or others adopt $E$ (deciding it may be a valid addition). Either way, by round $3$, all models match. The consensus answer would then include $A$, $B$, (and $E$ if it survived or is proven useful). Thus, the final answer is richer and more accurate than any single model’s initial output: it has all correct details combined and no erroneous detail $D$. This outcome is exactly what we want from consensus; maximize correct content, minimize incorrect content.
\end{itemize}
\section{Prototype Design} \label{sec:design}
In this section, we outline the prototype design in terms of system modules, data flow, and key algorithms. The design follows the theoretical approach but adds concrete specifications:
\subsection{System Modules}
\begin{itemize}
    \item \textbf{\textit{Query Handler}}: Accepts the user’s request (could be a question or task description) and initiates the consensus process. It could be an API endpoint itself that the user or application calls, and under the hood it triggers the multi-model consensus.
    \item \textbf{\textit{Model Interface Layer}}: Manages connections to each RM’s API. This layer handles sending prompts and receiving responses. It can also enforce timeouts or handle errors (if a model fails to respond, the system can retry or proceed without it, treating it as an offline node).
    \item \textbf{\textit{Consensus Controller}}: Implements the gossip rounds and checks for convergence. It stores the outputs from each model for each round in a data structure, e.g., a dictionary of the form $round\_outputs[r][i]$ = $O_i^r$. It also maintains a set of model IDs and can mark when each model has gotten each other’s output (for gossip tracking). The consensus controller is effectively the orchestrator mentioned in Section \ref{protodets}.
    \item \textbf{\textit{Prompt Generator}}: This component formulates the prompts sent to the models at each round. It takes as input the user query, the last answer of each model (if any), and a collection of peer answers, then produces a merged prompt string. As discussed, careful prompt design is crucial. This can be implemented as a template with slots for list of peer answers, etc., or even a small function that creates a comparison table in the prompt.
    \item \textbf{\textit{Comparer/Evaluator}}: After each round, this component compares the outputs. It could use string comparison or more advanced NLP techniques to judge equality of answers. In a sophisticated implementation, it could call another language model to ask: \textit{“Are these answers essentially the same?”} but that introduces an external dependency. Simpler approaches include normalized text matching (ignoring punctuation/casing) or computing an embedding similarity. This component decides whether to stop or continue.
    \item \textbf{\textit{Result Aggregator}}: Once consensus is reached, this module formats the final output. If slight differences remain between model outputs, it may choose the lexicographically first, or shortest, etc., or merge them if trivial. It could also attach a confidence score or an explanation of how consensus was reached (number of rounds, etc.).
\end{itemize}

\subsection{Data Flow}
\begin{itemize}
    \item \textbf{\textit{Initial Query}}: The \textit{Query Handler} receives the user’s query and forwards it to the \textit{Consensus Controller}. The controller calls the \textit{Model Interface} to send the query to all RMs simultaneously (round $0$). Responses are collected as $round\_outputs[0]$.
    \item \textbf{\textit{Gossip Rounds}}: For round $r = 1,2,...$ the controller uses the \textit{Prompt Generator} to create personalized prompts for each model $i$ that include all $O_j^{r-1}$ for $j \neq i$. These prompts are sent out via \textit{Model Interface} in parallel or sequence. As responses come back, the controller records them in $round\_outputs[r]$. Optionally, after each model responds, the controller could stream that answer to other models in real-time (more like actual gossip). But to keep it round-synchronized, we typically wait for all to respond, then consider that the completion of round $r$.
    \item \textbf{\textit{Evaluation}}: The \textit{Comparer/Evaluator} checks $round\_outputs[r]$ for consensus. If not achieved, increment $r$ and repeat. If some models failed to change when they perhaps should have (maybe one model stubbornly repeats its initial answer), the algorithm may decide to continue for a few rounds or possibly intervene (for example, if model $k$ has not changed at all while others converged, one might drop model $k$ out of the consensus or force it with a stronger prompt on next round). This is a design choice to handle non-cooperative behavior.
    \item \textbf{\textit{Termination}}: When consensus is detected or a maximum number of rounds is reached (to avoid infinite loops), the process stops. If consensus was reached properly, proceed to output. If the maximum rounds was hit without full agreement, as a fallback, the system could resort to majority vote or choose the answer that was most popular in last round as the final. However, the expectation is that with well-behaved models, consensus will usually be reached in a small number of rounds for factual queries.
    \item \textbf{\textit{Output}}: The \textit{Result Aggregator} prepares the final answer $O^{*}$, which the \textit{Query Handler} returns to the user. If needed, references or reasoning can be included (some 
    models might generate them during the process, which could be consolidated).
\end{itemize}
\subsection{Illustrative Pseudocode for Consensus Controller}
\begin{algorithm}
  \caption{Consensus Query for Multi-Model Reasoning}\label{alg:consensus}
  \begin{algorithmic}[1]
    \Procedure{ConsensusQuery}{$query,\,models$}
      \State $N \gets |models|$
      \State $round\_outputs[0] \gets \emptyset$
      \For{$i \gets 1$ \textbf{to} $N$}
        \State $round\_outputs[0][i]\gets models[i].\textsc{Ask}(query)$
      \EndFor
      \State $r \gets 1$
      \While{true}
        \For{$i \gets 1$ \textbf{to} $N$}
          \State $peerAnswers \gets \emptyset$
          \For{$j \gets 1$ \textbf{to} $N$}
            \If{$i \neq j$}
              \State \textsc{Append}($peerAnswers,\; (\text{$RM_j$}:round\_outputs[r-1][j])$)
            \EndIf
          \EndFor
          \State
            $prompt \gets$ 
            \texttt{"User question: }\,query\texttt{;}\\
            \quad\quad\quad\quad\quad\quad\quad\quad\quad\texttt{Peers (round }r-1\texttt{): }peerAnswers\texttt{;}\\
            \quad\quad\quad\quad\quad\quad\quad\quad\quad\texttt{Refine your answer accordingly."}
          \State $round\_outputs[r][i]\gets models[i].\textsc{Ask}(prompt)$
        \EndFor
        \State $answers \gets round\_outputs[r]$
        \If{\textsc{AllAnswersEquivalent}($answers$)}
          \State \Return \textsc{ChooseFinalAnswer}($answers$)
        \ElsIf{$r \ge \text{MAX\_ROUNDS}$}
          \State \Return \textsc{FallbackDecision}($answers$)
        \EndIf
        \State $r \gets r + 1$
      \EndWhile
    \EndProcedure
  \end{algorithmic}
\end{algorithm}

In the pseudocode described, $AllAnswersEquivalent$ implements the evaluator’s logic for consensus (e.g., all strings are identical or difference below the threshold), and $ChooseFinalAnswer$ can just pick $answers[0]$ since they are all equivalent. $FallbackDecision$ could choose the most frequent answer if the deliberation (discussion) among the models concluded without perfect agreement.

\textbf{\textit{Note on Scalability and Complexity}}: The prototype as described queries each model once per round, and includes the answers of other models in the prompt. This means that prompt size grows with $N$, and the computation per round for each model grows with $O(N)$ input size (since it reads all answers). For very large $N$, this is a concern, but in practice $N$ (the number of different foundation models) can be small (perhaps 3–10) because not many organizations produce these high-end RMs. The number of rounds needed is generally small too (we expect 1–3 in typical cases). Therefore, the overall overhead might be 2–5 queries per model for one user query, which is acceptable for high accuracy. If $N$ were large, an exact Hashgraph gossip (where each model only sees a subset at a time and the consensus is reached more gradually) could be employed to reduce input size per model. However, that would increase number of gossip steps. The trade-off between prompt size and number of rounds can be tuned.

\textbf{\textit{Security and Privacy Considerations}}: The output of each model is revealed to the others during gossip. If these models are from competing providers, sharing outputs might raise IP or data privacy issues. In a controlled ensemble (say all models are run by the same service or sharing is allowed), this is fine. If not, one could consider only sharing distilled information (e.g., not the full verbose answer but extracted facts or assertions) to limit exposure. Another approach could be to use an intermediate anonymized representation of content (to avoid revealing which model said what). These issues are outside the scope of our current design, but worth noting for real deployments.

\section{Evaluation Criteria} \label{sec:eval}
To rigorously evaluate the proposed consensus mechanism, we outline several criteria and experimental setups:
\begin{itemize}
    \item \textbf{\textit{Factual Accuracy Improvement}}: The primary metric is the correctness of the final (consensus) answer $O^{*}$ compared to the individual $O_i$ answers. This can be measured by testing the system on benchmark datasets of QA or reasoning tasks where ground truth is known. For each query, we check if $O^{*}$ is correct (fully or partially) and compare that to the correctness of each model on the same query. We expect to see that $O^{*}$ is at least as good as the answer from the best model, and often better than the answers from most models. An ideal outcome is that whenever at least one model knows the correct answer, the consensus answer does too. We can quantify how often consensus \textit{rescues} the correct answer from a minority model and how often it eliminates mistakes that a majority of models were making. A suitable benchmark can be a set of fact-based questions designed to trick some models into hallucinating (e.g., ambiguous or uncommon knowledge queries).

    \item \textbf{\textit{Hallucination Rate}}: We need to measure how many hallucinated or unsupported statements end up in the final output versus individual outputs. We can use human evaluators or existing factuality-checkers to label statements in the output of each model and in the consensus output as correct or hallucinated. The expectation is that the consensus output has a significantly lower rate of hallucinated statements. For example, if the answer from each model on average contains two false statements, the consensus answer can bring that down to near zero in ideal cases. We can also evaluate scenarios where all models hallucinate the same fact (consensus might not fix it because it appears consistent to them). This is a worst-case scenario for us. Measuring performance on those cases tells us the limitation; if every model is wrong in the same way, consensus will not help. However, the philosophy of our approach is that diversity in models means that it is less likely they all hallucinate identically. We can document any such failures.

    \item \textbf{\textit{Convergence Rounds and Stability}}: We need to track how many rounds it takes to reach consensus for each query. We aim for a low number (preferably 1 or 2 for most cases). If it often takes, say, 5+ rounds, that indicates either models are stubborn or oscillating. We can analyze any oscillation cases (where models keep changing their answers back and forth on alternating rounds), which might indicate a need to adjust the algorithm or prompt (like achieving coordination akin to making a binary consensus on a bit where oscillation is possible without proper tiebreaking). Ideally, we prove that our algorithm converges (which in theory, Hashgraph consensus would in a finite number of rounds for a given set of inputs). Observing stability: once converged, does the answer remain the same if another round were run? We may occasionally run one extra dummy round to verify nothing changes, confirming true consensus.

    \item \textbf{\textit{Content Preservation (Recall of correct info)}}: Another metric is how much correct information from the individual answers is preserved in the final answer. We do not want the consensus to strip away useful details just because they were initially unique. For this, if the ground truth answer contains certain key facts or steps, we check if these appear in the output of at least one model and if they appear in the final output. Ideally, the final output is a superset of the correct bits of each model's input (minus contradictions). This can be measured qualitatively or with overlap metrics. A high \textit{recall} of correct information means that the consensus is not losing information. Some loss may occur to achieve consistency, but we evaluate if that loss is acceptable (perhaps dropping a detail to avoid a contentious point).

    \item \textbf{\textit{Comparative Performance vs Baselines}}: We can compare our method against simpler baselines:
    \begin{itemize}
        \item \textbf{\textit{Majority Vote Output}}: If we just pick the answer most models gave initially (or pick one at random if all differ), how often is that right versus our final answer.
        \item \textbf{\textit{Best Individual Model}}: If we always relied on the model that is known to be best (e.g. if it generally has the highest accuracy), how does that fare versus consensus. This tests if consensus is adding value beyond the top model.
        \item \textbf{\textit{Chain-of-Thought self-consistency (single-model)}}: For tasks where one can query a model multiple times and vote, does using multiple different models with consensus do better?
    \end{itemize}
    These comparisons can highlight the value of cross-model consensus. Previous work suggests that unanimous agreement among models can be a strong indicator of correctness \citep{Pokharel2025}. We can examine if when our models all agree, they are almost always correct (which builds trust in using consensus answers).

    \item \textbf{\textit{Robustness to Malicious or Outlier Models}}: We can simulate a \textit{malicious} model that intentionally provides a wrong answer or extremely misleading content, and see if the consensus can still converge correctly. This ties to Byzantine tolerance. For example, with five models, if one always outputs a random incorrect answer, do the other four models override it and does it not pollute the final answer? Ideally yes; the bad actor’s influence should be neutralized by the majority of good actors. We can measure consensus accuracy as we vary the fraction of models that are adversarial or low-quality. Hashgraph theory says up to $1/3$ adversaries can be tolerated \citep{Baird2016swirlds}; we can test that empirically. If too many models are wrong, the final answer may be wrong. We want to identify that threshold in practice.

    \item \textbf{\textit{User Experience and Fidelity}}: Although harder to quantify, we want to ensure that the final answers are not only factually correct but also directly answer the request of the user with appropriate detail (fidelity to instructions). We can perform user studies or prompt evaluations to ensure that the consensus process does not produce overly conservative or vague answers. Sometimes, striving for consensus might lead to generic answers (the intersection of the knowledge from all models might be a safe but not very detailed answer). We can evaluate if that happens by comparing the level of detail of the final answer to the best individual answer. If we find a drop in detail or informativeness, we may consider ways to preserve detail (perhaps by including any non-conflicting additional details even if only one model had them).

    \item \textbf{\textit{Performance Metrics}}: Time and cost, i.e. how long and how many model calls does the process use on average. This is important for practical deployment. For expensive API calls, an approach that calls them many times may be costly. We can measure average consensus rounds and thus average calls. Also, does parallelization effectively keep latency roughly to one model call duration (since rounds can be parallelized)? If models are slow, multiple rounds will linearly increase latency, so there is a trade-off between thorough consensus and speed.
\end{itemize}

We aim to present results of experiments on a variety of question types: straightforward factual queries, ambiguous questions, multistep problems (e.g., math or logic puzzles where reasoning steps can be cross-checked), and even opinionated or open-ended queries (to see how consensus handles subjective content – possibly they converge to a blended perspective or highlight differing views). The evaluation can also include case studies to illustrate how a hallucination was eliminated or how a consensus answer was constructed from pieces of different answers.

\section{Implementation Challenges} \label{sec:challenge}
Implementing the system described above poses the following challenges:

\begin{itemize}
    \item \textbf{\textit{Prompt Design and Model Behavior}}: The biggest unknown is exactly how each black-box RM will respond when given the answers from its peers. We rely on cooperative behavior such that models will adjust their outputs toward correctness. There is a risk that a model may simply repeat its original answer regardless of input (it may trust itself more than others, especially if not instructed well). Models may also produce overly verbose or combined answers that are difficult to compare. We may need to iterate on prompt wording (engineering) to get the desired effect. Ensuring that models actually scrutinize peer answers (and not just blindly concatenate them or ignore them) is tricky. In some cases, models may get confused by seeing multiple answers (particularly if the answer from one peer is very wrong, the model might get distracted by it). We may need to include a prompt such as:\textit{“Some of the above may be incorrect. Only include what is correct in your answer.”}. Finding a general prompt that works across different API models is challenging, due to their different instruction-following styles. This may require empirical tuning.

    \item \textbf{\textit{Difference in Model Abilities}}: The models can have different levels of knowledge or reasoning. A highly capable model might dumb down its answer after seeing simpler answers from weaker models, or a weaker model might not fully comprehend a complex explanation from a stronger model. This asymmetry could slow convergence or result in a less detailed final answer (as noted, possibly the intersection of knowledge rather than the union). One mitigation is to allow models to provide justification or sources for their claims in the gossip, so that detailed reasoning from a stronger model can convince the others. However, integrating justifications increases prompt length and complexity. Alternatively, we could designate one round for clarifications: e.g., if a minority model insists on a different answer, have others ask it why (like a sub-dialogue). This starts to look like a full-blown debate, which is beyond our current scope. For now, we accept that some richness might be lost and consider it a trade-off for consensus; future work can explore preserving it.

    \item \textbf{\textit{Computational/Cost Overhead}}: Multiple rounds multiply the cost of using several large models. We have to optimize rounds where each model call costs tokens or latency. We may set a cutoff like three rounds to balance diminishing returns. There is also a state space explosion possibility: if models generate very long or divergent answers, sharing them in prompts can blow up token length. That could hit limits of context length. Techniques to compress information, such as summarizing the answers from peers or focusing on differences only, might be needed for very large outputs. Perhaps the orchestrator can identify the points of disagreement and only include those in prompts (\textit{“Model $1$ had a different figure for $X$ than others, who is correct?”}). This makes the process more efficient but requires NLP parsing of content to know what to focus on, thereby adding complexity.

    \item \textbf{\textit{Semantic Equivalence Checking}}: Deciding when answers are effectively the same is not trivial. If one answer says \textit{“The capital is Canberra, located in the Australian Capital Territory.”} and another says \textit{“Canberra.”}, they are equivalent on the core answer. We need an algorithm to detect this. Misidentification could either prematurely stop consensus (if we think they agreed but they did not on some detail) or prolong it (if we fail to acknowledge that they are basically saying the same thing). Using an AI judge model to compare can ironically introduce the opinion of another model. Alternatively, using embeddings similarity or simple heuristics (like check that all contain the same key terms/phrases for factual answers) may suffice. We must test and validate the equivalence criteria thoroughly, to avoid logical bugs in consensus detection.

    \item \textbf{\textit{Pathological Cases}}: It is possible the process does not converge. For example, imagine two models each trust their own answer and not the other’s. In round $1$, they swap answers but remain unconvinced and stick to their guns in round $2$; since nothing changed, they could go back and forth indefinitely. Our algorithm would then hit max rounds. We can incorporate a mechanism akin to tiebreakers in distributed consensus (like a predetermined priority or randomness to force a decision if stalemate). In blockchain consensus, usually if nodes keep disagreeing the protocol might stall, but in Hashgraph, the voting eventually yields a resolution because of the mathematical guarantee with enough rounds. For us, without a formal proof, we may sometimes need to force a resolution. One approach: if after several rounds two answers remain, have all models see an averaged or combined answer and vote between them explicitly (perhaps by asking each model to output just \textit{"$A$”} or \textit{“$B$”} corresponding to which answer they endorse). This would be an interesting fallback to implement and ensure a decision is made. Such complexities indicate this is not a foolproof system and requires careful handling of edge cases.

    \item \textbf{\textit{Malicious Model Behaviors}}: If a model deliberately outputs extremely misleading content, or tries to poison the consensus (e.g., by copying others but inserting slight inaccuracies to confuse), can the system cope? Hashgraph guarantees help if the majority are good, but a clever malicious model might adapt each round to continue causing disagreement (like always flip to a different answer than the majority to prevent unanimity). Our stopping rule of a max rounds would catch this and probably end up going with majority, but it shows consensus might not be reached if one agent is adversarial. We can decide to eject an agent if it appears to be the sole holdout across many rounds (assuming maybe it is faulty). However, identifying which model is the \textit{bad} one is non-trivial if you treat them equally. These are adversarial considerations that are a bit beyond normal expectations, but relevant if someone integrated, say, an untrusted open-source model into an ensemble of reliable ones.

    \item \textbf{\textit{Extension to Other Modalities or Multistep Plans}}: Our current design is for textual answer. For a task such as \textit{“produce a plan”}, consensus on the final output may require merging sequences, which adds complexity. As such, a challenge is how to generalize consensus beyond QA. One possibility is consensus on each step of a chain-of-thought (such as having models agree step by step). This complicates the procedure but may be more powerful. It remains an open challenge to apply consensus to arbitrary generative tasks, such as creative writing or design. As such, it is likely our current method is best suited for tasks with a notion of correctness.

    \item \textbf{\textit{Testing and Tuning}}: Given the novelty of combining these domains, a lot of testing is required to tune parameters: how strongly should the prompt push for agreement versus encourage independent thought, how many rounds to allow, how to weigh models if at all, etc. This challenge is more experimental, ensuring that our system is robust across various queries and not overfitted to a specific scenario.
\end{itemize}

Despite these challenges, we believe that each can be addressed incrementally. The core idea of gossiping outputs and virtually voting via model updates is flexible and can be iterated upon. The system can start with a simple implementation and incorporate more sophisticated techniques as needed. For example, adding a module to verify facts with an external database if the models cannot agree, which would combine retrieval-augmented generation (RAG) \citep{Lewis2020rag} with our approach. This essentially provides the tiebreaking to actual facts from external sources.

We also foresee that as AI models continue to improve and incorporate ways to introspect or verify their answers, our consensus approach would only become more effective. For instance, some new RMs can cite sources and the consensus could use that information. For example, choose the answer with strongest sources, etc., via virtual voting on confidence. Our framework is compatible with such enhancements.

\section{Conclusion} \label{sec:conclusion}
We have presented a novel approach to achieving reliable and high-fidelity outputs from multiple advanced reasoning models by borrowing concepts from distributed ledger consensus mechanisms. By treating each proprietary model as a node in a Hashgraph-like consensus network, our system enables collaborative verification of answers through gossip-based communication and virtual voting. This consensus mechanism addresses the key problem of hallucinations and inconsistent outputs in multi-LLM deployments: models cross-examine each other’s answers and converge on information that withstands collective scrutiny. We showed theoretically how the Hashgraph principles of gossip about gossip (rapid and full information dissemination) and Byzantine agreement (tolerating some faulty agents) translate to the domain of AI reasoning. Our proposed architecture demonstrates how such a consensus can be orchestrated in practice, guiding models through iterative refinement rounds.

The advantages of our approach over traditional ensembles are clear. Rather than simply taking a majority vote, which could ignore minority truths or be led astray by a unanimous falsehood, our mechanism encourages information sharing and correction. It maximizes the chance that correct information present in any model is retained and adopted by all, while spurious information is filtered out due to lack of support. In essence, it creates a \textit{discussion} among the models, with the Hashgraph consensus logic ensuring that the discussion yields a single agreed outcome efficiently and fairly. Our approach can be seen as a step toward a decentralized AI truth-validation system, where no single model is an oracle, but agreement emerges from the interplay of many. This is particularly powerful in industry settings where one may ensemble models from different vendors for critical tasks. Our consensus provides an additional layer of assurance on the results.

Through our exploration, we also identified several challenges and provided initial ideas to mitigate them, from prompt engineering to handling stubborn models. These highlight directions for future work. One important future direction is to formally analyze the convergence properties of the consensus algorithm in the context of language models. While we drew analogies to Hashgraph, the non-deterministic nature of language generation adds complexity. Another direction is testing on a wider range of tasks, including those requiring reasoning (e.g., math word problems) where models may carry out internal chain-of-thought; consensus could be applied at each reasoning step to ensure validity of the final answer. Additionally, integrating factual databases or tools in the loop could reinforce the consensus with external evidence (for example, if models disagree on a fact, the system could do a quick web lookup as an arbitrator).

In conclusion, the proposed consensus mechanism is a promising framework for harnessing the collective intelligence of multiple AI models while minimizing the impact of their individual flaws. It embodies a fusion of multi-agent systems and distributed consensus theory, pointing to a future where AI systems can verify themselves in a decentralized manner before presenting information to humans. We expect this concept to inspire further research in both the AI and blockchain communities: the former to enhance AI reliability and the latter to find new applications of consensus algorithms beyond financial ledgers, in the realm of knowledge and truth aggregation. By demonstrating feasibility through our prototype design and analysis, we hope to lay the groundwork for real-world implementations of multi-model consensus, ultimately leading to AI assistants that are not only powerful but also trustworthy and aligned with factual reality.

Ultimately, minimizing hallucinations in AI is about ensuring that truth prevails over randomness or error. Our Hashgraph-inspired consensus mechanism is a significant step in that direction, leveraging diversity for robustness; an approach very much in the spirit of both democratic decision-making and fault-tolerant computing.

\bibliographystyle{unsrtnat}
\bibliography{references}

\begin{thebibliography}{16}
\providecommand{\natexlab}[1]{#1}
\providecommand{\url}[1]{\texttt{#1}}
\expandafter\ifx\csname urlstyle\endcsname\relax
  \providecommand{\doi}[1]{doi: #1}\else
  \providecommand{\doi}{doi: \begingroup \urlstyle{rm}\Url}\fi

\bibitem[Huang et~al.(2023)Huang, Yu, Ma, Zhong, Feng, Wang, Chen, Peng, Feng, Qin, and Liu]{Huang2023}
Lei Huang, Weijiang Yu, Weitao Ma, Weihong Zhong, Zhangyin Feng, Haotian Wang, Qianglong Chen, Weihua Peng, Xiaocheng Feng, Bing Qin, and Ting Liu.
\newblock A survey on hallucination in large language models: Principles, taxonomy, challenges, and open questions.
\newblock \emph{ACM Transactions on Information Systems}, 2023.
\newblock \doi{10.1145/3703155}.
\newblock URL \url{https://arxiv.org/pdf/2311.05232}.

\bibitem[Dursun and \"{U}st\"{u}ndağ(2021)]{Dursun2021}
Taner Dursun and Burak~Berk \"{U}st\"{u}ndağ.
\newblock A novel framework for policy based on-chain governance of blockchain networks.
\newblock \emph{Information Processing \& Management}, 58\penalty0 (4):\penalty0 102556, July 2021.
\newblock ISSN 0306-4573.
\newblock \doi{10.1016/j.ipm.2021.102556}.
\newblock URL \url{http://dx.doi.org/10.1016/j.ipm.2021.102556}.

\bibitem[Baird(2016)]{Baird2016swirlds}
Leemon Baird.
\newblock The swirlds hashgraph consensus algorithm: Fair, fast, byzantine fault tolerance.
\newblock Technical Report SWIRLDS-TR-2016-01, Swirlds, Inc., 2016.
\newblock URL \url{https://www.swirlds.com/downloads/SWIRLDS-TR-2016-01.pdf}.
\newblock Accessed: 2025-05-02.

\bibitem[Crary(2021)]{Crary2021}
Karl Crary.
\newblock Verifying the hashgraph consensus algorithm, 2021.
\newblock URL \url{https://arxiv.org/pdf/2102.01167}.

\bibitem[Hedera(2025{\natexlab{a}})]{Hedera2025gossip}
Hedera.
\newblock Gossip about gossip, 2025{\natexlab{a}}.
\newblock URL \url{https://docs.hedera.com/hedera/core-concepts/hashgraph-consensus-algorithms/gossip-about-gossip}.
\newblock Accessed: 2025-05-02.

\bibitem[Morgan(2025)]{morgan2025selfcheckgpt}
Abby Morgan.
\newblock Selfcheckgpt for llm evaluation, 2025.
\newblock URL \url{https://www.comet.com/site/blog/selfcheckgpt-for-llm-evaluation/}.
\newblock Accessed: 2025-05-02.

\bibitem[Zheng et~al.(2023)Zheng, Chiang, Sheng, Zhuang, Wu, Zhuang, Lin, Li, Li, Xing, Zhang, Gonzalez, and Stoica]{Zheng2023}
Lianmin Zheng, Wei-Lin Chiang, Ying Sheng, Siyuan Zhuang, Zhanghao Wu, Yonghao Zhuang, Zi~Lin, Zhuohan Li, Dacheng Li, Eric~P. Xing, Hao Zhang, Joseph~E. Gonzalez, and Ion Stoica.
\newblock Judging llm-as-a-judge with mt-bench and chatbot arena, 2023.
\newblock URL \url{https://arxiv.org/pdf/2306.05685}.

\bibitem[Wang et~al.(2022)Wang, Wei, Schuurmans, Le, Chi, Narang, Chowdhery, and Zhou]{Wang2022}
Xuezhi Wang, Jason Wei, Dale Schuurmans, Quoc Le, Ed~Chi, Sharan Narang, Aakanksha Chowdhery, and Denny Zhou.
\newblock Self-consistency improves chain of thought reasoning in language models, 2022.
\newblock URL \url{https://arxiv.org/pdf/2203.11171}.

\bibitem[Lamport(2001)]{Lamport2001paxos}
Leslie Lamport.
\newblock Paxos made simple, 2001.
\newblock URL \url{https://lamport.azurewebsites.net/pubs/paxos-simple.pdf}.
\newblock Accessed: 2025-05-02.

\bibitem[Ongaro and Ousterhout(2014)]{Ongaro2014raft}
Diego Ongaro and John Ousterhout.
\newblock In search of an understandable consensus algorithm, 2014.
\newblock URL \url{https://raft.github.io/raft.pdf}.
\newblock Accessed: 2025-05-02.

\bibitem[Castro and Liskov(1999)]{Castro1999paxos}
Miguel Castro and Barbara Liskov.
\newblock Practical byzantine fault tolerance.
\newblock In \emph{Proceedings of the Third Symposium on Operating Systems Design and Implementation (OSDI)}, pages 173--186, New Orleans, LA, 1999. USENIX Association.
\newblock URL \url{https://pmg.csail.mit.edu/papers/osdi99.pdf}.
\newblock Accessed: 2025-05-02.

\bibitem[Irving et~al.(2018)Irving, Christiano, and Amodei]{Irving2018debate}
Geoffrey Irving, Paul Christiano, and Dario Amodei.
\newblock Ai safety via debate.
\newblock 2018.
\newblock URL \url{https://arxiv.org/pdf/1805.00899}.
\newblock Accessed: 2025-05-02.

\bibitem[Ogunsina and DeLaurentis(2022)]{Ogunsina2022}
Kolawole Ogunsina and Daniel DeLaurentis.
\newblock Enabling integration and interaction for decentralized artificial intelligence in airline disruption management.
\newblock \emph{Engineering Applications of Artificial Intelligence}, 109:\penalty0 104600, 2022.
\newblock \doi{10.1016/j.engappai.2021.104600}.
\newblock URL \url{https://doi.org/10.1016/j.engappai.2021.104600}.
\newblock Accessed: 2025-05-02.

\bibitem[Pokharel et~al.(2025)Pokharel, Dantu, Zaman, Talapuru, and Quach]{Pokharel2025}
Apurba Pokharel, Ram Dantu, Shakila Zaman, Sirisha Talapuru, and Vinh Quach.
\newblock Achieving unanimous consensus in decision making using multi-agents, 2025.
\newblock URL \url{https://arxiv.org/pdf/2504.02128}.

\bibitem[Hedera(2025{\natexlab{b}})]{Hedera2025virtualvoting}
Hedera.
\newblock Virtual voting, 2025{\natexlab{b}}.
\newblock URL \url{https://docs.hedera.com/hedera/core-concepts/hashgraph-consensus-algorithms/virtual-voting}.
\newblock Accessed: 2025-05-02.

\bibitem[Lewis et~al.(2020)Lewis, Perez, Piktus, Petroni, Karpukhin, Goyal, Küttler, Lewis, tau Yih, Rocktäschel, Riedel, and Kiela]{Lewis2020rag}
Patrick Lewis, Ethan Perez, Aleksandra Piktus, Fabio Petroni, Vladimir Karpukhin, Naman Goyal, Heinrich Küttler, Mike Lewis, Wen tau Yih, Tim Rocktäschel, Sebastian Riedel, and Douwe Kiela.
\newblock Retrieval-augmented generation for knowledge-intensive nlp tasks.
\newblock \emph{arXiv preprint arXiv:2005.11401}, 2020.
\newblock URL \url{https://arxiv.org/pdf/2005.11401}.
\newblock Accessed: 2025-05-02.

\end{thebibliography}

\end{document}